\documentclass[letterpaper, 10 pt, conference]{ieeeconf}

\IEEEoverridecommandlockouts

\usepackage{url}
\usepackage{cite}
\usepackage{amsmath}
\usepackage{amssymb}
\usepackage{graphicx}

\begin{document}

\title{\LARGE \bf 
    Eva-Tracker: ESDF-update-free, Visibility-aware Planning \\ with Target Reacquisition for Robust Aerial Tracking
}

\author{
    Yue Lin, Yang Liu*, Dong Wang, Huchuan Lu
    \thanks{All authors are with Dalian University of Technology, Dalian 116024, China. *Corresponding author: Yang Liu, \tt\small{ly@dlut.edu.cn}.}
}

\maketitle
\thispagestyle{empty}
\pagestyle{empty}

\begin{abstract}

The Euclidean Signed Distance Field (ESDF) is widely used in visibility evaluation to prevent occlusions and collisions during tracking. However, frequent ESDF updates introduce considerable computational overhead. To address this issue, we propose Eva-Tracker, a visibility-aware trajectory planning framework for aerial tracking that eliminates ESDF updates and incorporates a recovery-capable path generation method for target reacquisition. First, we design a target trajectory prediction method and a visibility-aware initial path generation algorithm that maintain an appropriate observation distance, avoid occlusions, and enable rapid replanning to reacquire the target when it is lost. Then, we propose the Field of View ESDF (FoV-ESDF), a precomputed ESDF tailored to the tracker's field of view, enabling rapid visibility evaluation without requiring updates. Finally, we optimize the trajectory using differentiable FoV-ESDF-based objectives to ensure continuous visibility throughout the tracking process. Extensive simulations and real-world experiments demonstrate that our approach delivers more robust tracking results with lower computational effort than existing state-of-the-art methods. The source code is available at \url{https://github.com/Yue-0/Eva-Tracker}.

\end{abstract}

\section{Introduction}

Maintaining visibility is essential for aerial autonomous tracking robots to continuously observe their targets. In cluttered environments, trackers must avoid collisions, prevent occlusions, maintain an appropriate tracking distance and observation angle, and promptly recover when the target is lost. Consequently, developing an efficient and robust tracking framework for aerial robots that ensures continuous visibility, handles target loss, and operates within limited onboard computational resources remains a significant challenge.

Numerous existing methods~\cite{ji2022elastic, han2021fast, zhang2023auto, ren2024intention, lee2024bpmp, zou2025reacter} prevent collisions by constraining the tracker within safe corridors. However, they often struggle to maintain target visibility in 3D environments, leading to frequent occlusions or suboptimal observation distances. The most widely adopted approaches~\cite{zou2025reacter, jeon2019online, jeon2020integrated, wang2021visibility, lin2024safety, gao2024adaptive, zhang2025roar} that jointly consider visibility and safety build on the Euclidean Signed Distance Field (ESDF). They leverage gradients of the ESDF to generate occlusion-free and collision-free trajectories. However, updating the ESDF is computationally intensive, which hinders real-time performance. Additionally, trajectory optimization typically covers only a small subregion of the ESDF update range, resulting in redundant computations~\cite{zhou2020ego}. More importantly, these methods frequently rely on complex multi-objective optimization, making it challenging to balance occlusion avoidance with maintaining an appropriate tracking distance.

A promising approach to reducing the computational burden of ESDF-based methods is to utilize pre-built ESDFs, such as the RC-ESDF~\cite{geng2023robo}, which is specifically designed for collision assessment in robots with arbitrary shapes. By tailoring the ESDF to the robot's geometry in advance, this method eliminates the need for updates. However, it primarily focuses on collision avoidance and does not incorporate visibility, which is crucial for aerial tracking tasks.

\begin{figure}[t]
    \centering
    \vspace{1 mm}
    \includegraphics[width=\linewidth]{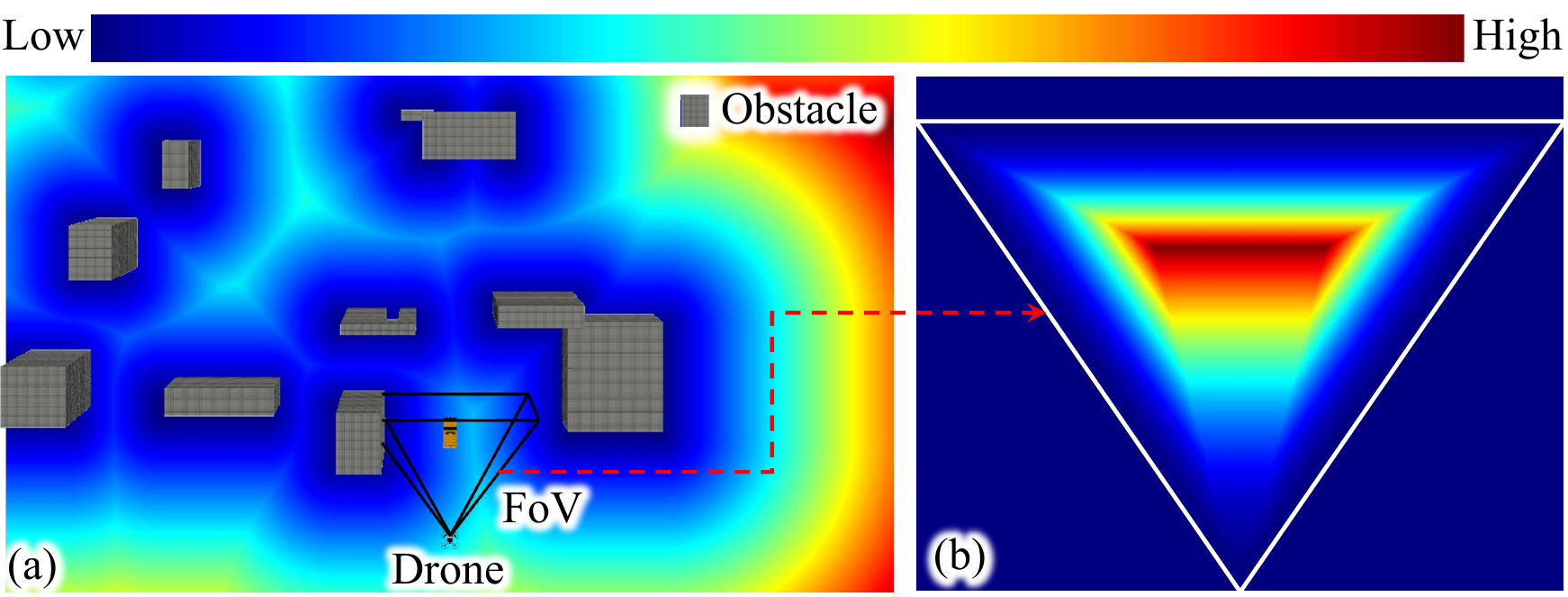}
    \caption{Comparison between the traditional Euclidean Signed Distance Field (ESDF) and the proposed Field of View ESDF (FoV-ESDF). (a) The traditional ESDF encodes the distance from each voxel to its nearest obstacle across a wide spatial range. It must be updated in real time to reflect environmental changes, resulting in high computational overhead and substantial memory usage. (b) A slice of the FoV-ESDF, with white lines indicating the boundaries of the tracker's FoV. The FoV-ESDF only encodes distances from voxels within the FoV to their nearest FoV boundary, while ignoring voxels outside the FoV. It is constructed once, requires significantly less memory, and eliminates the need for subsequent updates.}
    \vspace{-3 mm}
    \label{fig:esdf}
\end{figure}

Inspired by the concept of pre-built ESDFs, we propose the Field of View ESDF (FoV-ESDF), a novel ESDF specifically designed for visibility in tracking scenarios, as illustrated in Fig.~\ref{fig:esdf}. Unlike the RC-ESDF, which focuses solely on collision avoidance, the FoV-ESDF explicitly models target visibility within the tracker's FoV. It assigns gradients to obstacles located within the FoV, pushing them outward to preserve an unobstructed view, while ignoring those outside the FoV. Additionally, it generates a gradient for the target to maintain it centered and at an appropriate tracking distance, thereby ensuring continuous visibility. This visibility-aware design not only substantially reduces computational overhead but also unifies the optimization objectives of observation angle alignment, distance regulation, and occlusion avoidance through the FoV-ESDF, thereby significantly enhancing tracking robustness in cluttered 3D environments.

\begin{figure*}[t]
    \centering
    \vspace{3 mm}
    \includegraphics[width=\linewidth]{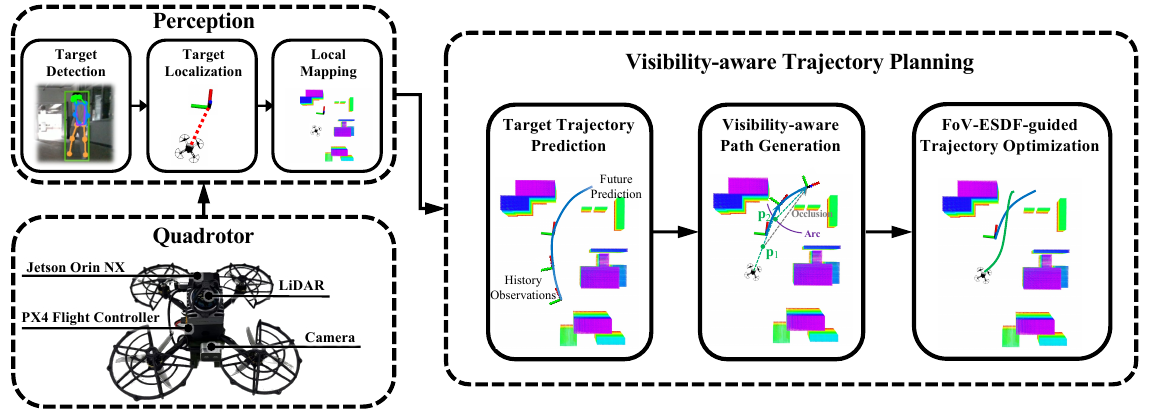}
    \caption{Overview of the proposed system and the real-world drone platform. Left: The perception module processes sensor data for target detection, localization, and local mapping, all of which are executed entirely on the drone's onboard computing device. Right: The trajectory planning module comprises three core components: target trajectory prediction, visibility-aware initial path generation, and FoV-ESDF-guided trajectory optimization. These components collaborate to produce occlusion-free tracking trajectories, ensuring continuous target visibility and enabling recovery when the target is lost.}
    \label{fig:framework}
    \vspace{-3 mm}
\end{figure*}

Initial path generation is another critical component of aerial tracking, as it provides an initial feasible solution for subsequent trajectory optimization. Many existing methods~\cite{ji2022elastic, ren2024intention, lin2024safety} rely on global searches, which incur high computational costs. Moreover, these methods fail immediately when the target is lost, lacking recovery mechanisms. To address these issues, we propose an efficient visibility-aware initial path generation algorithm that directly samples candidate observation points based on the predicted target position and the tracker's pose, followed by visibility-aware adjustments to ensure an occlusion-free path. When the target is lost, the algorithm continues to generate paths using predicted positions, enabling prompt reacquisition. Compared to traditional methods, our algorithm significantly reduces computation time and enhances tracking robustness.

Building upon the FoV-ESDF and the initial path generation algorithm, we propose \textbf{Eva}-Tracker, an \textbf{E}SDF-update-free, \textbf{V}isibility-\textbf{A}ware trajectory planning framework for aerial tracking. As illustrated in Fig.~\ref{fig:framework}, our system is composed of three tightly integrated components: target trajectory prediction, visibility-aware initial path generation, and FoV-ESDF-guided trajectory optimization. The prediction module generates smooth and continuous future target trajectories based on historical observations. The path generation module rapidly constructs occlusion-free initial paths that maintain an appropriate observation distance and support target reacquisition when the target is lost. Finally, the trajectory optimization module refines the tracking trajectory by maximizing visibility through a set of differentiable FoV-ESDF-based cost functions. Together, these modules enable robust and real-time aerial tracking in complex environments.

To evaluate the performance of our approach, we benchmark Eva-Tracker against several state-of-the-art tracking methods~\cite{wang2021visibility, ji2022elastic, lin2024safety}. The experimental results demonstrate that Eva-Tracker achieves more robust tracking while significantly reducing computational costs. Furthermore, we deploy Eva-Tracker on a real-world quadrotor system and conduct tests in both indoor and outdoor environments. The results confirm that our method enables stable tracking even in challenging scenarios, highlighting its effectiveness and practicality in real-world applications.

In summary, the major contributions of this paper are:

\begin{itemize}
    \item An efficient visibility-aware initial path generation algorithm with tracking recovery capability.
    \item A pre-built FoV-ESDF enables efficient visibility evaluation without requiring updates.
    \item A series of differentiable objective functions based on the FoV-ESDF to optimize the tracker's pose.
    \item Comprehensive simulations and real-world experiments demonstrate superior tracking performance while reducing computational costs.
\end{itemize}

\section{Related Work}

Recently, numerous learning-based trackers~\cite{bhagat2020uav, pan2021fast, masnavi2022visibility, yan2023long, masnavi2024differentiable, lu2025yopov2} have demonstrated impressive performance across various tracking tasks. However, they typically necessitate extensive training or environment-specific fine-tuning, and often generalize poorly to unfamiliar environments. Although FAN~\cite{maalouf2024follow} improves robustness by integrating several large-scale models, its substantial computational overhead limits its applicability to real-time deployment on resource-constrained platforms.

To enhance adaptability in unknown environments and facilitate deployment on edge devices, a variety of modular methods have been explored. Fast-Tracker~\cite{han2021fast} combines a kinematic search front end with a trajectory optimization back end. However, it neglects visibility, often resulting in overly close tracking distances or target loss. SF-Tracker~\cite{lin2024safety} maintains visibility by minimizing the observation angle error, but it does not account for occlusions. Elastic Tracker~\cite{ji2022elastic} models the visible area using 2D sectors, overlooking occlusions caused by 3D obstacles. In contrast, our method leverages the FoV-ESDF to simultaneously account for occlusions and observation distance in 3D environments, thereby ensuring continuous target visibility.

A widely adopted approach involves leveraging the ESDF for visibility optimization to avoid occlusions. The receding-horizon methods~\cite{jeon2019online, jeon2020integrated} mitigate occlusion by maximizing the ESDF values along the line from the tracker to the target. The visibility-aware tracking methods~\cite{wang2021visibility, gao2024adaptive} approximate the FoV using multiple spheres and maximize ESDF values of these spheres to prevent occlusion. However, constructing and updating the ESDF is time-consuming and computationally expensive. Moreover, these methods require additional objective functions to separately optimize tracking distance and observation angle. In contrast, our method employs a precomputed FoV-ESDF, eliminating the need for updates and enabling the joint optimization of occlusion, tracking distance, and observation angle within a unified framework.

Some studies have investigated target reacquisition. Fast-Tracker~\cite{han2021fast, pan2021fast} determines the navigation goal by predicting the target's future trajectory but does not consider visibility. ROAR~\cite{zhang2025roar} employs a Markov chain-based reacquisition mechanism. However, it estimates target motion solely from the current observation, which can result in significant prediction errors. In contrast, our method predicts the target's trajectory using a sequence of historical observations and explicitly incorporates visibility constraints.

\section{Visibility-aware Path Generation}

\subsection{Target Trajectory Prediction}

Given an observation time interval $\tau$ and a sequence of $m + 1$ historical observation positions $\mathbf{h}_0, \mathbf{h}_1, \cdots, \mathbf{h}_m \in \mathbb{R}^3$, we model the target's trajectory using a Bézier curve
\begin{equation}
    \boldsymbol{\rho}(t) = \sum_{i=0}^{n} \dfrac{n! (T+t)^i (T-t)^{n-i}}{i! (n-i)! (2T)^n} \boldsymbol{\varrho}_i, \ t \in [-T, T],
\end{equation}
where $n = 2m$, $T = m\tau$, and $\boldsymbol{\varrho}_i \in \mathbb{R}^3$ represents the control point of the Bézier curve. The past observations correspond to $t \le 0$, while future predictions correspond to $t > 0$. 

To obtain the predicted trajectory, we solve the following constrained optimization problem:
\begin{equation}
    \begin{aligned}
        \underset{\boldsymbol{\varrho}}{\min} & \quad \int_{-T}^{T} \Vert \boldsymbol{\rho}^{(2)}(t) \Vert^2 \mathrm{d}t, \\
        \mathrm{s.t.} & \quad \boldsymbol{\rho}(i\tau - T) = \mathbf{h}_i, \ i \in \lbrace0, 1, \cdots, m\rbrace.
    \end{aligned}
\end{equation}
Smoothness is ensured by minimizing the high-order derivatives of the trajectory, while equality constraints enforce that the predicted trajectory passes through historical observation points. This problem can be reformulated as an equality-constrained convex quadratic programming (EQP) problem, which admits a closed-form solution~\cite{nocedal2006numerical}.

\subsection{Visibility-aware Initial Path Generation}

\begin{figure}[t]
    \vspace{3 mm}
    \centering
    \includegraphics[width=\linewidth]{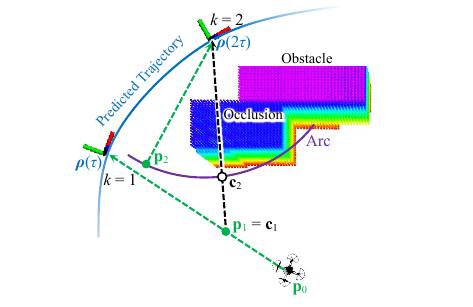}
    \caption{Illustration of visibility-aware initial path generation.}
    \label{fig:path}
    \vspace{-3 mm}
\end{figure}

To generate a suitable initial path for subsequent optimization, we propose a visibility-aware path generation algorithm that accounts for both observation distance and potential occlusions. This algorithm produces a sequence of waypoints $\mathbf{p}_1, \mathbf{p}_2, \cdots, \mathbf{p}_m \in \mathbb{R}^3$, as illustrated in Fig.~\ref{fig:path}.

Given the predicted target position $\boldsymbol{\rho}(k\tau)$ and the tracker's previous position $\mathbf{p}_{k-1}$, we first enforce a preferred observation distance $d$. Specifically, we draw a ray from the target toward the tracker and select a candidate point $\mathbf{c}_k$ along this ray at a distance $d$ from the target:
\begin{equation}
    \mathbf{c}_k = \boldsymbol{\rho}(k\tau) - d\dfrac{\mathbf{\Pi}_2 \left( \boldsymbol{\rho}(k\tau) - \mathbf{p}_{k-1} \right)}{\Vert \mathbf{\Pi}_2 (\boldsymbol{\rho}(k\tau) - \mathbf{p}_{k-1}) \Vert},
\end{equation}
where $\mathbf{\Pi}_2 = (\mathbf{e}_1, \mathbf{e}_2, \mathbf{0})$ is a projection matrix that extracts the horizontal component, and $\mathbf{e}_i$ represents the $i$-th column of the third-order identity matrix.

Next, we check for occlusions along the line connecting $\mathbf{c}_k$ and $\boldsymbol{\rho}(k\tau)$. As shown in Fig.~\ref{fig:path} for $k = 1$, if no occlusion is detected, $\mathbf{c}_k$ is directly assigned as the waypoint $\mathbf{p}_k$. Conversely, as illustrated in Fig.~\ref{fig:path} for $k = 2$, if an occlusion is detected, we generate a circular arc around $\boldsymbol{\rho}(k\tau)$ with a radius $d$, represented as the purple arc in Fig.~\ref{fig:path}. Starting from $\mathbf{c}_k$, we perform a bidirectional search along the arc to identify the nearest occlusion-free point, which is then selected as $\mathbf{p}_k$. This process ensures that each waypoint maintains a clear line of sight to the target while preserving the desired observation distance, thereby providing a robust initial path for subsequent trajectory optimization. 

In addition, this algorithm inherently enables target reacquisition. Even when the target is temporarily undetected, the system persists in generating occlusion-free waypoints at the optimal observation distance relative to the predicted target position. Once the target reappears within the FoV, the tracker seamlessly resumes normal tracking without the need for additional recovery mechanisms.

\section{FoV-ESDF-guided Trajectory Optimization}

\subsection{FoV-ESDF}

In this section, we detail the FoV-ESDF, which is precomputed offline based on the tracker's FoV, and does not require updates. Furthermore, it employs on-demand occlusion evaluation, concentrating solely on obstacles within the tracker's FoV, thereby significantly reducing computational overhead.

As illustrated by the black pyramid in Fig.~\ref{fig:esdf}(a), we model the FoV as a quadrilateral pyramid, denoted as
\begin{equation}
    \left \lbrace (x, y, z) \ | \ 0 < x \le D, \dfrac{\left \vert y \right \vert}{x} \le \tan \dfrac{\alpha}{2}, \dfrac{\left \vert z \right \vert}{x} \le \tan \dfrac{\beta}{2} \right \rbrace,
\end{equation}
where $(x, y, z)$ denotes the coordinate in the tracker's frame, $\alpha$ and $\beta$ represent the horizontal and vertical FoV angles of the camera, and $D$ corresponds to the optimal observation distance $d$. Specifically, $D$ is selected such that the point at $(d, 0, 0)$ achieves the maximum FoV-ESDF value.

For points within the FoV, the FoV-ESDF value is defined as the Euclidean distance to the nearest FoV boundary, while for points outside the FoV, it is set to zero. A planar cross-section of the FoV-ESDF is shown in Fig.~\ref{fig:esdf}(b). We construct the FoV-ESDF using the distance transform algorithm~\cite{felzenszwalb2012distance}, and both its value and gradient at any point can be obtained through cubic linear interpolation~\cite{zhou2019robust}. Despite being precomputed, the FoV-ESDF supports dynamic obstacle reasoning by transforming obstacle positions to the tracker's frame at each time step, enabling efficient visibility optimization.

\begin{figure}[t]
    \vspace{3 mm}
    \centering
    \includegraphics[width=\linewidth]{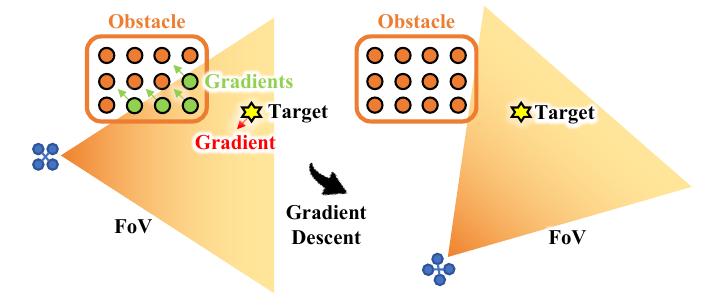}
    \caption{Illustration of FoV-ESDF-guided visibility optimization. Green circles represent obstacles within the FoV, while green arrows indicate the direction of the gradient descent generated by the FoV-ESDF for these obstacles. A red arrow indicates the direction of the gradient descent generated by the FoV-ESDF for the target. Orange circles represent obstacles that do not enter the FoV, which are excluded from the calculation.}
    \label{fig:optimization}
    \vspace{-3 mm}
\end{figure}

To illustrate the optimization process of FoV-ESDF for visibility, we present an example in Fig.~\ref{fig:optimization}. Initially, some obstacles obstruct the target. The FoV-ESDF computes gradients for both the target and these obstacles, which are then propagated to the tracker's trajectory. The trajectory is subsequently optimized using gradient descent to maximize target visibility while minimizing occlusion. Specifically, the gradients generated for the obstacles are used to push them away from the FoV, while the gradients generated for the target are employed to maintain the target at the center of the FoV. After optimization, the obstacles are no longer within the FoV, and the target appears centered in the camera view.

\subsection{Trajectory Representation and Problem Formulation}

We parameterize the tracker's waypoints and yaw angles into a MINCO trajectory~\cite{wang2022geometrically}, formulated as
\begin{equation}
    \mathfrak{T} = \left \lbrace \mathbf{p}(t), \psi(t) \ | \ [\mathbf{C}, \boldsymbol{\Theta}] = \mathcal{M}(\mathbf{P}, \boldsymbol{\Psi}, \mathbf{t})\right \rbrace,
\end{equation}
where $\mathbf{p}(t): \mathbb{R}_+ \mapsto \mathbb{R}^3$ and $\psi(t): \mathbb{R}_+ \mapsto \mathbb{R}$ are polynomials of order $2s - 1$ with $m$ segments, $\mathbf{P} = [\mathbf{p}_1, \mathbf{p}_2, \cdots, \mathbf{p}_m] \in \mathbb{R}^{3 \times m}$ denotes the waypoints, $\boldsymbol{\Psi} \in \mathbb{R}^{1 \times m}$ represents the yaw angles at these waypoints, the vector $\mathbf{t} = [t_1, t_2, \cdots, t_m] \in \mathbb{R}_+^m$ specifies the duration of each trajectory segment, the tensor $\mathbf{C} = [\mathbf{C}_1, \mathbf{C}_2, \cdots, \mathbf{C}_m] \in \mathbb{R}^{2s \times 3 \times m}$ and the matrix $\boldsymbol{\Theta} = [\boldsymbol{\theta}_1, \boldsymbol{\theta}_2, \cdots, \boldsymbol{\theta}_m] \in \mathbb{R}^{2s \times m}$ contain the coefficients for the trajectory segments. The function $\mathcal{M}$ maps $\mathbf{P}$, $\boldsymbol{\Psi}$ and $\mathbf{t}$ to $\mathbf{C}$ and $\boldsymbol{\Theta}$ with linear complexity, allowing any second-order continuous cost function $J(\mathbf{C}, \boldsymbol{\Theta}, \mathbf{t})$ to have a gradient applicable to the MINCO trajectory~\cite{wang2022geometrically}. The $i$-th segment of $\mathbf{p}$ and $\psi$ is denoted as
\begin{equation}
    \mathbf{p}_i(t) = \mathbf{C}_i^\top \boldsymbol{\beta}(t), \quad \psi_i(t) = \boldsymbol{\theta}_i^\top \boldsymbol{\beta}(t),
\end{equation}
where $\boldsymbol{\beta}(t) = [1, t, t^2, \cdots, t^{2s-1}]^\top$ represents the natural basis. The trajectory optimization problem is formulated as
\begin{equation}
    \underset{\mathbf{P}, \boldsymbol{\Psi}, \mathbf{t}}{\min} J_v(\mathbf{C}, \boldsymbol{\Theta}, \mathbf{t}) + J_e(\mathbf{C}, \boldsymbol{\Theta}, \mathbf{t}) + \lambda_p J_p(\mathbf{C}, \boldsymbol{\Theta}, \mathbf{t}),
\end{equation}
where $J_v$ represents the visibility cost, $J_e$ is the energy cost, $J_p$ is the penalty term, with $\lambda_p$ serving as its weighting factor. We solve this problem using L-BFGS algorithm~\cite{liu1989limited} with Lewis-Overton line search~\cite{lewis2013nonsmooth}.

\subsection{Visibility Cost}

\subsubsection{Overview}
The visibility cost is designed to ensure optimal observation by minimizing occlusion while maintaining a desirable observation angle and distance. It consists of three components, formulated as
\begin{equation}
    J_v = \lambda_o \mathcal{J}_o + \lambda_v \mathcal{J}_v + \lambda_a \mathcal{J}_a,
\end{equation}
where $\mathcal{J}_o$ penalizes occlusions, $\mathcal{J}_v$ enforces an optimal observation pose, and $\mathcal{J}_a$ regulates the observation angle to facilitate convergence. The hyperparameters $\lambda_o$, $\lambda_v$, and $\lambda_a$ control the relative importance of each term. Each component follows a unified numerical integration formulation:
\begin{equation}
    \mathcal{J}_\mu = \sum_{i=1}^{m} \dfrac{t_i}{k_i} \sum_{j=0}^{k_i} \eta_j f_\mu \left( \mathbf{p}_i, \psi_i, \boldsymbol{\rho}_i, \dfrac{jt_i}{k_i} \right), \mu \in \lbrace o, v, a \rbrace,
\end{equation}
where $\left( \eta_0, \eta_1, \eta_2, \cdots, \eta_{k_i-1}, \eta_{k_i} \right) = (0.5, 1, 1, \cdots, 1, 0.5)$, $k_i$ represents the number of samples for the $i$-th numerical integration, and $\boldsymbol{\rho}_i$ denotes the target trajectory segment:
\begin{equation}
    \boldsymbol{\rho}_i(t) = \boldsymbol{\rho} \left( \max \left\lbrace T, t + \sum_{k=1}^{i-1} t_k\right\rbrace \right).
\end{equation}
The gradients of the visibility cost are computed as
\begin{subequations}
    \begin{equation}
        \dfrac{\partial \mathcal{J}_\mu}{\partial t_i} = \dfrac{1}{k_i} \sum_{j=0}^{k_i} \eta_j \left( f_\mu + \dfrac{jt_i}{k_i} \dfrac{\partial f_\mu}{\partial t} \right),
    \end{equation}
    \begin{equation}
        \dfrac{\partial \mathcal{J}_\mu}{\partial \mathbf{C}_i} = \dfrac{t_i}{k_i} \sum_{j=0}^{k_i} \eta_j \dfrac{\partial f_\mu}{\partial \mathbf{C}_i}, \ \dfrac{\partial \mathcal{J}_\mu}{\partial \boldsymbol{\theta}_i} = \dfrac{t_i}{k_i} \sum_{j=0}^{k_i} \eta_j \dfrac{\partial f_\mu}{\partial \boldsymbol{\theta}_i}.
    \end{equation}
\end{subequations}

\subsubsection{Occlusion Penalty}
The occlusion penalty is designed to prevent obstacles from entering the FoV by penalizing their presence. It is defined as
\begin{subequations}
    \begin{equation}
        f_o \left( \mathbf{p}_i, \psi_i, \boldsymbol{\rho}_i, t \right) = \dfrac{1}{2} \left( \sum_{k=1}^{N_t} \Xi(\mathbf{v}_{k,t}) \right)^2,
    \end{equation}
    \begin{equation}
        \mathbf{v}_{k,t} = \mathbf{R}(\psi_i(t))(\mathbf{w}_{k,t} - \mathbf{p}_i(t)),
    \end{equation}
\end{subequations}
where $\Xi: \mathbb{R}^3 \mapsto \mathbb{R}$ represents the FoV-ESDF value, $N_t$ is the number of obstacle points inside the FoV at time $t$, $\mathbf{w}_{k,t}$ is the $k$-th obstacle point in the world frame, $\mathbf{v}_{k,t}$ represents the coordinate of this point in the tracker's frame, and
\begin{equation}
    \mathbf{R}(\theta) = \begin{bmatrix}
        \cos \theta & \sin \theta & 0 \\
        -\sin \theta & \cos \theta & 0 \\
        0 & 0 & 1
    \end{bmatrix}.
\end{equation}
Note that $\Xi(\mathbf{v}_{k,t})$ is precomputed offline, while the obstacle point $\mathbf{w}_{k,t}$ is updated online to handle dynamic occlusions.

Let $w = \sqrt{2f_o}$, and $\mathbf{d}_{k,t} = \mathbf{R}'(\psi_i(t)) [\mathbf{w}_{k,t} - \mathbf{p}_i(t)]$, where $\mathbf{R}'(\theta)$ represents the derivative of the rotation matrix $\mathbf{R}(\theta)$. The gradients of the occlusion penalty $f_o$ are computed as 
\begin{subequations}
    \begin{equation}
        \dfrac{\partial f_o}{\partial \boldsymbol{\theta}_i} = w\sum_{k=1}^{N_t} \boldsymbol{\beta}(t) \mathbf{d}_{k,t}^\top \nabla \Xi(\mathbf{v}_{k,t}),
    \end{equation}
    \begin{equation}
        \dfrac{\partial f_o}{\partial \mathbf{C}_i} = -w\sum_{k=1}^{N_t} \boldsymbol{\beta}(t) \nabla^\top \Xi(\mathbf{v}_{k,t}) \mathbf{R}(\psi_i(t)),
    \end{equation}
    \begin{equation}
        \dfrac{\partial f_o}{\partial t} = w\sum_{k=1}^{N_t} \left[ \mathbf{d}_{k,t} - \mathbf{R}(\psi_i(t))\mathbf{p}_i'(t) \right]^\top \nabla \Xi(\mathbf{v}_{k,t}),
    \end{equation}
\end{subequations}
where $\nabla \Xi(\mathbf{v}_{k,t})$ represents the gradient of the FoV-ESDF at $\mathbf{v}_{k,t}$, which can be calculated by cubic linear interpolation.

\subsubsection{Observation Penalties}
To ensure optimal visibility, we optimize the tracker's observation pose by maximizing the target's FoV-ESDF value, formulated as
\begin{equation}
    f_v \left( \mathbf{p}_i, \psi_i, \boldsymbol{\rho}_i, t \right) = \dfrac{1}{2} \left[ \Xi(d\mathbf{e}_1) - \Xi \left( \mathbf{R}(\psi_i(t)) \boldsymbol{\delta}_i(t) \right) \right]^2,
\end{equation}
where $\boldsymbol{\delta}_i(t) = \boldsymbol{\rho}_i(t) - \mathbf{p}_i(t)$ represents the position of the target relative to the tracker, and $\Xi(d\mathbf{e}_1)$ is the maximum FoV-ESDF value. The gradient of this penalty can be computed in the same manner as $f_o$.

As shown in Fig.~\ref{fig:esdf}(b), due to the 3D nature of the FoV, the point at which the maximum value of the FoV-ESDF is obtained is not unique on a horizontal plane. To ensure that the target remains centered in the FoV, the optimal yaw angle of the tracker should be
\begin{equation}
    \varphi_i(t) = \mathrm{atan2} \left( \mathbf{e}_2^\top \boldsymbol{\delta}_i(t), \mathbf{e}_1^\top \boldsymbol{\delta}_i(t) \right).
\end{equation}

To enforce this constraint, we design a straightforward angular penalty
\begin{equation}
    f_a \left( \mathbf{p}_i, \psi_i, \boldsymbol{\rho}_i, t \right) = 1 - \cos \delta_i(t),
\end{equation}
where $\delta_i(t) = \psi_i(t) - \varphi_i(t)$ represents the angle of the target relative to the tracker. The gradient of $f_a$ is given by
\begin{subequations}
    \begin{equation}
        \dfrac{\partial f_a}{\partial \boldsymbol{\theta}_i} = \boldsymbol{\beta}(t) \sin \delta_i(t),
    \end{equation}
    \begin{equation}
        \dfrac{\partial f_a}{\partial \mathbf{C}_i} = \dfrac{\sin \delta_i(t)}{\Vert \mathbf{\Pi}_2 \boldsymbol{\delta}_i(t) \Vert^2} \boldsymbol{\beta}(t) \boldsymbol{\delta}_i^\top(t) (-\mathbf{e}_2, \mathbf{e}_1, \mathbf{0}),
    \end{equation}
    \begin{equation}
        \dfrac{\partial f_a}{\partial t} = \sin \delta_i(t) \left[ \psi_i'(t) - \dfrac{\mathbf{e}_3^\top (\boldsymbol{\delta}_i(t) \times \boldsymbol{\delta}_i'(t))}{\Vert \mathbf{\Pi}_2 \boldsymbol{\delta}_i(t) \Vert^2} \right].
    \end{equation}
\end{subequations}

\subsection{Energy Cost and Penalty Functions}

To maintain a smooth target view and minimize motion blur, we select $s = 3$, corresponding to jerk minimization~\cite{ji2022elastic}. Additionally, we incorporate a temporal regularization term to synchronize the duration of the tracker's trajectory with that of the target. Thus, the energy cost is formulated as $J_e =$
\begin{equation}
    \sum_{i=1}^{m} \int_{0}^{t_i} \left[ \left\Vert \mathbf{p}_i^{(s)} \right\Vert^2 + \left( \psi_i^{(s)} \right)^2 \right] \mathrm{d}t + \gamma \left( T - \sum_{i=1}^{m} t_i \right)^2,
\end{equation}
where $\gamma$ is the weight of the regularization term. The value and gradient of $J_e$ can be computed analytically~\cite{wang2022geometrically}.

To ensure safety and kinematic feasibility, we introduce a numerical integration penalty term, formulated as
\begin{subequations}
    \begin{equation}
        J_p = \sum_{i=1}^{m} \dfrac{t_i}{k_i} \sum_{j=0}^{k_i} \eta_j \sum_{\mu \in P} G_\mu \left( \mathbf{p}_i, \psi_i, \dfrac{jt_i}{k_i} \right),
    \end{equation}
    \begin{equation}
        G_d \left( \mathbf{p}_i, \psi_i, t \right) = H \left( \mathbf{p}_i(t), \psi_i(t) \right),
    \end{equation}
    \begin{equation}
        G_{p_h} \left( \mathbf{p}_i, \psi_i, t \right) = \max \left\lbrace \left\Vert \mathbf{p}_i^{(h)}(t) \right\Vert^2 - v_h^2, 0 \right\rbrace, 
    \end{equation}
    \begin{equation}
        G_{a_h} \left( \mathbf{p}_i, \psi_i, t \right) = \max \left\lbrace \left( \psi_i^{(h)}(t) \right)^2 - \omega_h^2, 0 \right\rbrace,
    \end{equation}
\end{subequations}
where $h \in \lbrace 1, 2 \rbrace$, and $P = \lbrace d, p_1, p_2, a_1, a_2 \rbrace$. $G_d$ enforces safety constraints, the function $H$ is a differentiable collision metric derived from RC-ESDF, with further details available in~\cite{geng2023robo}. The hyperparameters $v_1$, $v_2$, $\omega_1$, and $\omega_2$ represent the maximum velocity, acceleration, angular velocity, and angular acceleration of the tracker, respectively.

\section{Simulations and Benchmarking}

\subsection{Implementation Details}

We benchmark Eva-Tracker against Elastic Tracker~\cite{ji2022elastic}, Vis-Planner~\cite{wang2021visibility}, and SF-Tracker~\cite{lin2024safety}. These open-source methods represent state-of-the-art tracking performance. All trackers are configured with an identical FoV of $\alpha \times \beta = 69.4^\circ \times 42.5^\circ$, and the optimal observation distance is set to $d = 2.5$m. The maximum linear and angular velocities of the tracker are set to $2$m/s and $1.5$rad/s, respectively. During the experiment, the target's position is continuously transmitted to the tracker. The hyper-parameters of Eva-Tracker are set as follows: $\tau = 0.5$s, $m = 2$, $\gamma = \lambda_p = 10^5$, $\lambda_o = 100$, $\lambda_d = 10^3$, and $\lambda_a = 10^4$.  All simulations are conducted on an Intel NUC Phantom Canyon equipped with an Intel Core i7-1165G7 CPU and an NVIDIA GeForce RTX 2060 GPU.

\subsection{Benchmark Comparisons}

\begin{figure}[b]
    \centering
    \includegraphics[width=\linewidth]{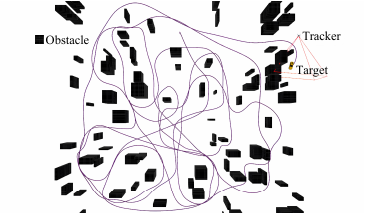}
    \caption{The random simulation environment and the target's trajectory.}
    \label{fig:env}
\end{figure}

As shown in Fig.~\ref{fig:env}, we construct a simulated environment with 100 randomly placed obstacles to evaluate the generality and robustness of the four methods. The target follows a 200-second trajectory involving sharp turns, movement toward the tracker, and traversal through cluttered regions.

We evaluate tracking performance of four methods using four metrics summarized in Tab.~\ref{tab:metric}: tracking distance (TD), yaw angle error (AE), occlusion rate (OR), and failure rate (FR). A tracking failure is defined as the target becomes occluded or leaves the tracker's FoV. SF-Tracker exhibits frequent failures owing to its lack of optimization for occlusion and tracking distance. Vis-Planner experiences suboptimal performance as a consequence of complex visibility-aware multi-objective optimization. Although Elastic Tracker effectively maintains accurate yaw through 2D sectors, it struggles with managing 3D occlusions. In contrast, Eva-Tracker consistently upholds the desired observation distance and achieves the lowest occlusion rate, thereby yielding the highest success rate among all evaluated methods.

\begin{table}[t]
    \vspace{3 mm}
    \centering
    \caption{Tracking Performance Comparison}
    \begin{tabular}{c c c c c}
        \hline
        Method & TD    (m)       & AE    (rad)     & OR(\%)  & FR(\%) \\
        \hline
        SF-Tracker~\cite{lin2024safety}       
               & 2.89 $\pm$ 0.59 & 0.14 $\pm$ 0.13 & 27.50   & 40.58  \\
        Vis-Planner~\cite{wang2021visibility} 
               & 2.87 $\pm$ 0.58 & 0.11 $\pm$ 0.08 & 17.32   & 28.50  \\
        Elastic tracker~\cite{ji2022elastic}  
               & 2.52 $\pm$ 0.41 & \textbf{0.09 $\pm$ 0.08}
                                                   & 8.71    & 12.27  \\
        \textbf{Eva-Tracker}          
               & \textbf{2.50 $\pm$ 0.24} 
                                 & 0.10 $\pm$ 0.08 & \textbf{4.36} 
                                                             & \textbf{6.75}  \\
        \hline
    \end{tabular}
    \label{tab:metric}
    \vspace{-3 mm}
\end{table}

\begin{figure}[hb]
    \vspace{-3 mm}
    \centering
    \includegraphics[width=\linewidth]{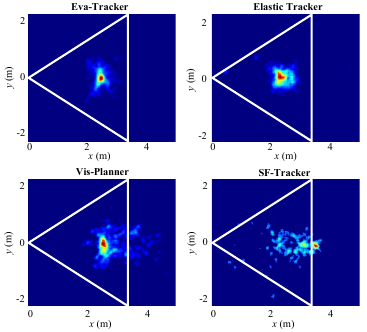}
    \caption{Comparison of the distribution heatmap of the target position relative to the tracker's FoV on the $x$-$y$ plane. The heatmaps illustrate the frequency of the target appearing at various locations, with a more concentrated distribution near the center indicating stable visibility. Warmer colors denote a higher frequency of the target at corresponding locations, and white triangles mark the boundaries of the tracker's FoV.}
    \label{fig:heatmap}
\end{figure}

To further analyze visibility, we present the spatial distribution of the target’s projection onto the tracker's FoV in the $x$-$y$ plane in Fig.~\ref{fig:heatmap}. SF-Tracker lacks explicit optimization for tracking distance, exhibits a large standard deviation in tracking distance. Vis-Planner struggles to balance multiple visibility-aware objective functions, resulting in a scattered distribution. Both our method and Elastic Tracker maintain a compact and centered distribution. However, unlike Elastic Tracker, our approach explicitly accounts for 3D occlusions, thereby further enhancing visibility robustness.

We evaluate the average computation time of these methods during tracking, as summarized in Tab.~\ref{tab:time}. The proposed visibility-aware initial path generation algorithm is an order of magnitude faster than the others, producing feasible paths in significantly less time. Regarding trajectory optimization, Elastic Tracker achieves the fastest speed due to its simplified 2D-sector-based formulation, which omits 3D occlusion reasoning. In comparison, our Eva-Tracker fully accounts for visibility in 3D space and achieves the fastest optimization speed among ESDF-based approaches, such as Vis-Planner and SF-Tracker, as it does not require ESDF updates.

\begin{table}[t]
    \vspace{3 mm}
    \centering
    \caption{Comparison of Average Computation Time}
    \begin{tabular}{c c c c}
        \hline
        Method                                & Path (ms)       & Optimize (ms)    & Total (ms) \\
        \hline
        Elastic tracker~\cite{ji2022elastic}  & 0.56 $\pm$ 0.19 & 6.56  $\pm$ 0.92 & 7.12       \\
        Vis-Planner~\cite{wang2021visibility} & 0.79 $\pm$ 0.78 & 35.94 $\pm$ 4.81 & 36.69      \\
        SF-Tracker~\cite{lin2024safety}       & 0.93 $\pm$ 0.46 & 36.59 $\pm$ 5.60 & 37.52      \\
        \textbf{Eva-Tracker}                  & 0.02 $\pm$ 0.01 & 10.73 $\pm$ 3.89 & 10.75      \\
        \hline
    \end{tabular}
    \label{tab:time}
    \vspace{-3 mm}
\end{table}

\subsection{Visibility-aware Tracking in a 3D Scene} 

\begin{figure}[hb]
    \centering
    \includegraphics[width=\linewidth]{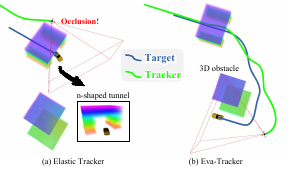}
    \caption{Comparison of tracking trajectories in a 3D structured environment. (a) Elastic Tracker loses visibility when the target enters the n-shaped tunnel. (b) Eva-Tracker maintains clear visibility throughout, including while the target traverses the tunnel and navigates around a two-layer obstacle.}
    \label{fig:tracking3d}
    \vspace{-2 mm}
\end{figure}

To effectively demonstrate Eva-Tracker's capability to maintain visibility in intricate 3D environments, we design a representative scenario in which the target traverses an n-shaped tunnel and subsequently navigates around a two-layer 3D obstacle. The tracking trajectories of both Eva-Tracker and Elastic Tracker~\cite{ji2022elastic} are illustrated in Fig.~\ref{fig:tracking3d}.

The Elastic Tracker loses sight of the target within the tunnel, leading to tracking failure. In contrast, the Eva-Tracker utilizes FoV-ESDF to analyze 3D occlusions, ensuring continuous visibility throughout the process and demonstrating its superior ability to handle 3D structures.

\section{Real-world Experiments}

\begin{figure*}[t]
    \vspace{3 mm}
    \centering
    \includegraphics[width=\linewidth]{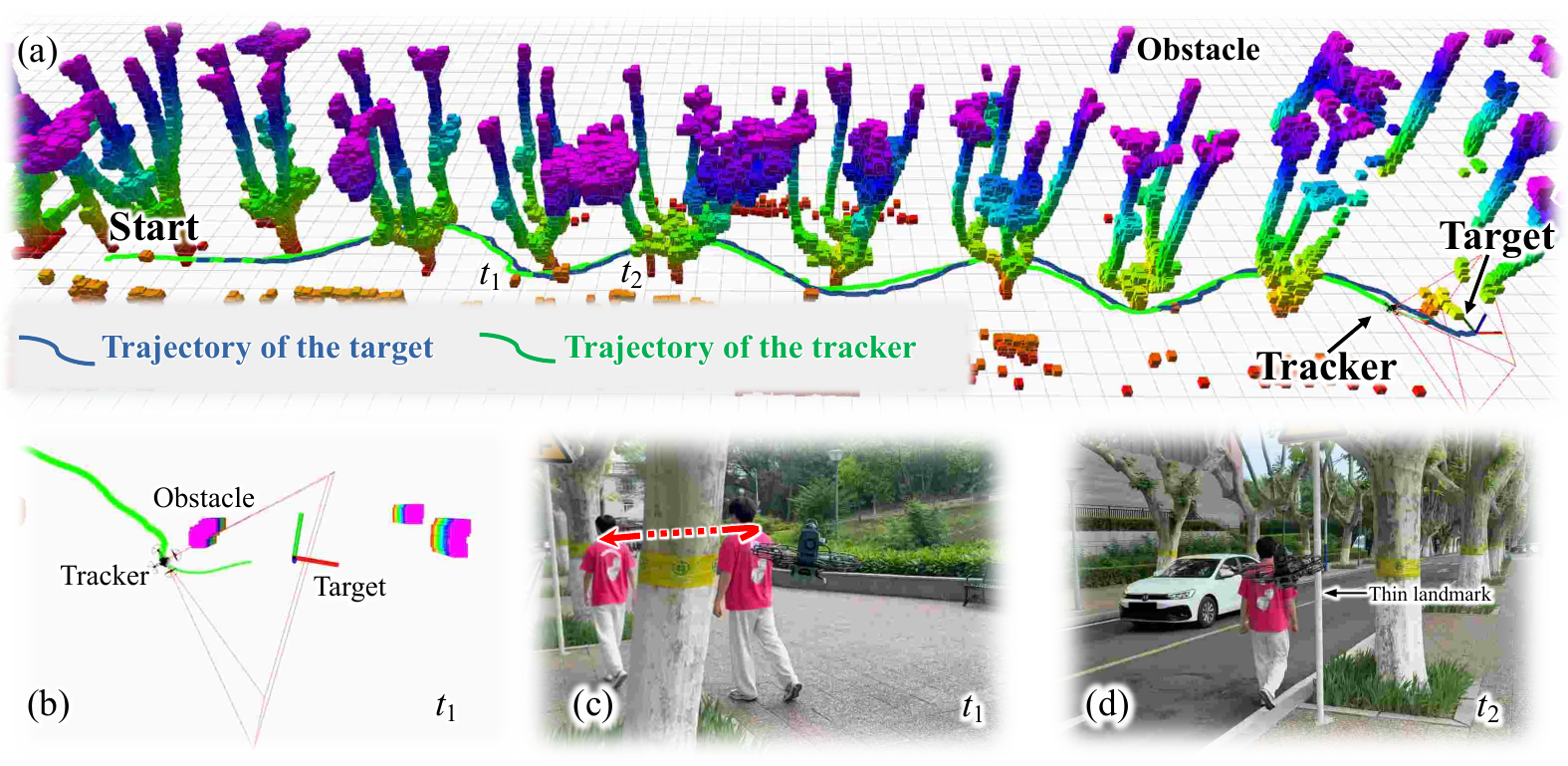}
    \caption{Illustration of outdoor experiment. (a) Trajectories of the tracker and the target. (b) The planned trajectory of the tracker and the position of the target at time $t_1$. (c) The tracker and the target at time $t_1$. (d) The tracker and the target at time $t_2$.}
    \label{fig:outdoor}
    \vspace{-3 mm}
\end{figure*}

To validate the practicality of Eva-Tracker, we deploy it on a real-world quadrotor platform, as shown in Fig.~\ref{fig:framework}. The quadrotor is equipped with a Livox Mid-360 LiDAR for environmental perception and a camera for target detection. All computations are performed onboard using an NVIDIA Jetson Orin NX.

\begin{figure}[t]
    \vspace{2 mm}
    \centering
    \includegraphics[width=\linewidth]{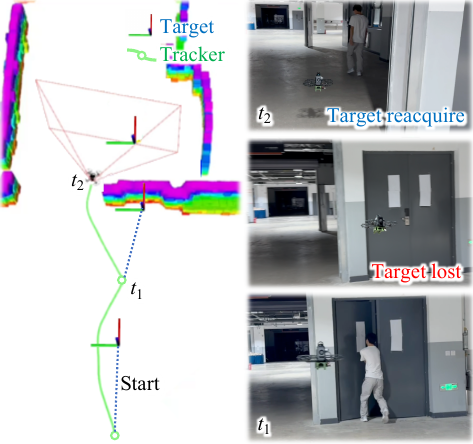}
    \caption{Illustration of indoor experiment. The target walks through a door at time $t_1$, after which the door closes and occludes the tracker’s view. The tracker proactively navigates around the obstacle, successfully reacquires the target at time $t_2$, and resumes tracking.}
    \label{fig:indoor}
    \vspace{-5 mm}
\end{figure}

We adopt Faster-LIO~\cite{bai2022faster} for self-localization. The target is a person, whose pose is inferred by first detecting skeletal keypoints using YOLOv11~\cite{khanam2024yolov11}, followed by 3D pose estimation via MonoLoco~\cite{bertoni2019monoloco}. When visual detection fails, the predicted position is used to attempt target reacquisition. To distinguish the target from obstacles, we remove LiDAR point clouds located within the estimated target radius prior to mapping and trajectory planning.

We evaluate Eva-Tracker in an unfamiliar outdoor environment populated with multiple trees. As shown in Fig.~\ref{fig:outdoor}, the target repeatedly moves around tree trunks, which can easily cause occlusions. Throughout the experiment, the tracker proactively maneuvers around obstacles while maintaining an unobstructed line of sight to the target. The flight trajectory exhibits occlusion-free, collision-free, and smooth motion, with the tracker maintaining a stable tracking distance from the target. Notably, at time $t_1$, the target suddenly changes direction after circumventing a large tree trunk. The tracker promptly adjusts its trajectory to avoid target loss. Additionally, at time $t_2$, the tracker is able to adjust its trajectory to avoid occlusion even when the target passes by a thin landmark. This experiment demonstrates that Eva-Tracker can reliably maintain both visibility and safety during tracking.

To verify the robustness of Eva-Tracker in handling temporary target loss, we conduct an experiment in an unfamiliar, dynamic indoor environment, as shown in Fig.~\ref{fig:indoor}. The target walks through a narrow doorway, and the door subsequently closes, completely blocking the tracker's line of sight. At time $t_1$, the tracker uses the target's estimated trajectory from historical observations to generate occlusion-free waypoints at the optimal observation distance relative to the predicted positions. This enables the quadrotor to navigate around the obstacle without hesitation. Upon reaching the other side of the doorway at time $t_2$, the target reappears within the FoV, indicating successful tracking recovery, and the system seamlessly transitions back to normal tracking without requiring explicit recovery commands. The resulting trajectory confirms that Eva-Tracker can maintain operational continuity and rapidly reacquire the target in complex, dynamic environments where visibility is temporarily lost.

\section{Conclusion and Limitations}

In this paper, we propose Eva-Tracker, an ESDF-update-free, visibility-aware planning framework for aerial tracking that integrates a recovery-capable path generation mechanism. By leveraging a precomputed FoV-ESDF, Eva-Tracker eliminates the computational overhead associated with frequent ESDF updates. Combined with an efficient visibility-aware initial path generation algorithm, Eva-Tracker enables real-time performance and robust visibility maintenance in challenging 3D environments. Even during temporary target loss, Eva-Tracker facilitates rapid tracking recovery without requiring additional recovery procedures. Extensive simulations and real-world experiments demonstrate that Eva-Tracker outperforms state-of-the-art methods in both visibility preservation and recovery robustness in 3D scenarios.

Despite its advantages, Eva-Tracker has two primary limitations. First, the target reacquisition mechanism is effective only for short-term losses. If the target’s actual trajectory deviates significantly from the predicted trajectory during prolonged occlusion, successful recovery becomes difficult. Additionally, the FoV-ESDF relies on a fixed camera FoV assumption, which limits its applicability to platforms equipped with dynamic FoV cameras such as zoom lenses or gimbal-mounted systems. Future work will address these limitations by developing a long-horizon motion prediction algorithm to manage extended target loss and by exploring adaptive FoV models to accommodate dynamic fields of view.

\bibliographystyle{IEEEtran}
\bibliography{root}

\end{document}